%% file: egpaper_final.tex
\documentclass[10pt,twocolumn,letterpaper]{article}

\usepackage{cvpr}
\usepackage{times}
\usepackage{epsfig}
\usepackage{graphicx}
\usepackage{amsmath}
\usepackage{amssymb}
\usepackage[table,xcdraw]{xcolor}
\usepackage{adjustbox}
\usepackage{multirow}
\usepackage{floatrow}
\usepackage{booktabs}
\usepackage{float}

\usepackage[breaklinks=true,bookmarks=false]{hyperref}
\DeclareMathOperator{\EX}{\mathbb{E}}

\cvprfinalcopy 

\def\cvprPaperID{12} 

\usepackage{xcolor} 
\usepackage{soul} 
\usepackage{xspace}
\usepackage{amsmath}

\usepackage{amssymb}
\usepackage{pifont}

\usepackage{mwe}
\usepackage{cvpr}
\usepackage{blindtext}
\usepackage{graphicx}
\usepackage{caption}

\newcommand{\cmark}{\ding{51}}%
\newcommand{\xmark}{\ding{55}}%

\usepackage{amsmath}

\definecolor{Green}{HTML}{AFDAAF}

\begin{document}

\title{PoliTO-IIT-CINI Submission to the EPIC-KITCHENS-100 Unsupervised Domain Adaptation Challenge for Action Recognition}

\author{Mirco Planamente$^{1,2,3}$ \quad
Gabriele Goletto $^{1}$ \quad 
Gabriele Trivigno $^{1}$ \quad
Giuseppe Averta $^{1}$ \quad
Barbara Caputo\textsuperscript{1,3} \\

\and \textsuperscript{1} Politecnico di Torino\\
{\tt\small {name.surname}@polito.it}

\and \textsuperscript{2} Istituto Italiano di Tecnologia\\
{\tt\small {name.surname}@iit.it}

\and \textsuperscript{2} Consortium Cini\\
}

\maketitle

\input{root/abstract}
\date{
\begin{figure*}
    \centering
    \includegraphics[width=1.0\linewidth]{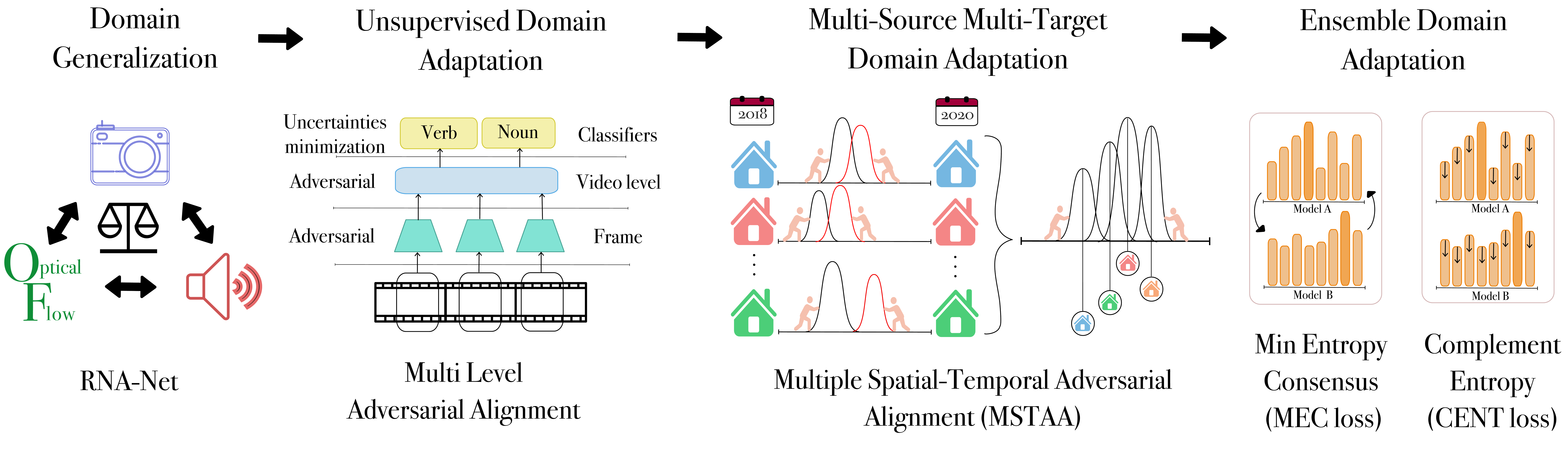}
    \caption{An overview of the proposed approach. It can be summarized in four main aspects: \textbf{1.} Domain Generalization through RNA-Net \cite{planamente2022domain}, \textbf{2.} Unsupervised Domain Adaptation via Multi-Level Adversarial Alignment and entropy minimization, \textbf{3.} Multi-Source Multi-Target Domain Adaptation extension and \textbf{4.} Ensemble Domain Adaptation losses.}
    \label{fig:teaser}
\end{figure*}}
\input{root/Intro}
\input{root/DG_DA}

\input{root/backbone}

\input{root/Exp}

{\small
\bibliographystyle{ieee_fullname}
\bibliography{egbib}
}

\end{document}

%% file: root/abstract.tex
\begin{abstract}

In this report, we describe the technical details of our submission to the EPIC-Kitchens-100 Unsupervised Domain Adaptation (UDA) Challenge in Action Recognition. 
To tackle the domain-shift which exists under the UDA setting, we first exploited a 
recent Domain Generalization (DG) technique, called
Relative Norm Alignment (RNA). 
Secondly, we extended this approach to work on unlabelled target data, enabling a simpler adaptation of the model to the target distribution in an unsupervised fashion. To this purpose, we included in our framework UDA algorithms, such as 
multi-level adversarial alignment and attentive entropy. By analyzing the challenge setting, we notice the presence of a secondary concurrence shift in the data, which is usually called environmental bias. It is caused by the existence of different environments, i.e., kitchens. To deal with these two shifts (environmental and temporal), we extended our system to perform Multi-Source Multi-Target Domain Adaptation. Finally, we employed distinct models in our final proposal to leverage the potential of popular video architectures, and we introduced two more losses for the ensemble adaptation. Our submission (entry ‘plnet') is visible on the leaderboard and ranked in 2nd position for \textit{‘verb’}, and in 3rd position for both \textit{‘noun’} and \textit{‘action’}.

\end{abstract}

%% file: root/Intro.tex
\section{Introduction}

First person action recognition offers a wide range of opportunities and challenges, thanks to the use of wearable devices to capture the current state of the user and of the environment. Very often, indeed, the actions of the subject are captured through a video-camera placed on the head of the user. As a consequence, in contrast with most CV tasks, the major feature of this scenario is that source data is intrinsically characterized by rich multi-modal information, thanks to the proximity of the sensor to the action scene. As a result, sensor fusion between visual and auditory cues can be a powerful method to fully exploit the knowledge available in the data. However, the particular setup of data collection also comes with several difficulties: i) ego-motion represents a significant source of noise for the dataset, because changes in head posture cause a shift in the point-of-view and  background. While from one side this effect can be exploited as an intrinsic attention mechanism, it may also introduce confusion between ego-motion and the real action of the subject. An approach to mitigate this effect could be to complement RGB data with other motion-related sources, such as the optical flow;  ii) model predictions tend to be strongly correlated with the surrounding environment, which represents a bias in the dataset (usually referred to as \emph{environmental bias}), thus resulting in decreased performances when the environment changes (e.g. different kitchens).   
%
%
%
%
%
%
In this report, we discuss the idea that, to fully exploit the potential of data sources, and to mitigate the performances drop across domains, it is crucial to properly combine several sensing modalities, including audio, video, and motion. This is particularly true for cross-domain scenarios, where test data are extracted from a different distribution w.r.t. the training data (i.e. different users and/or kitchens). 
Indeed, the effect of domain shift is not consistent across different sensing modalities, and some of them may suffer in some cases where others are more robust. 

The reason is that domain shifts are not all of the same nature. For instance, the optical flow is more focused on the motion in the scene, rather than appearance, and is therefore less sensitive to environmental changes, thus showing higher robustness than the visual modality when changing environment~\cite{munro2020multi}. On the other side, the domain shift of auditory information is very different from the visual one (e.g., the sound of ‘cut' will differ from a plastic to a wooden cutting board). For all those reasons, the classifier should be able to assess - depending on the conditions - which modality is more informative, and therefore should be considered more for the final prediction.


To this purpose, authors of \cite{planamente2022domain} recently proposed a multi-modal framework, called Relative Norm Alignment network (RNA-Net), which aims at progressively aligning the feature norms of audio and visual (RGB) modalities among multiple sources in a Domain Generalization (DG) setting, where target data are not available during training. Interestingly, the authors showed that \emph{merely feeding all the source domains to the network without applying any adaptive techniques leads to sub-optimal performance, while a multi-source domain alignment allows the network to promote domain-agnostic features. }

Including all the aforementioned considerations, we developed the method adopted in the challenge with the following steps (see also Figure \ref{fig:teaser}):
\begin{enumerate}
    \item RNA-Net was extended to the Flow modality, obtaining remarkable results without accessing target data;
    \item with further modifications, RNA-Net was adapted to work with unlabelled target data under the standard Unsupervised Domain Adaptation (UDA) setting;
    \item the challenge's setting was revisited by identifying a new concurrent shift denominated "environmental bias". Our framework was modified accordingly to perform Multi-Source Multi-Target Domain Adaptation; 
    \item the final submission was obtained by combining different model streams by means of  DA-based losses, namely Min-Entropy Consistency (MEC) and Complement Entropy (CENT).
\end{enumerate}





%% file: root/DG_DA.tex
\section{Our Approach}\label{OurApp}
In this section, we first describe the DG approach used. Then, we show our UDA framework and its extension for Multi-Source Multi-Target Domain Adaptation. Finally, we demonstrate how to re-define existing DA-based losses to induce consistency between different architectures.
\subsection{Domain Generalization}
The multi-source nature of the proposed challenge setting makes it perfect to deal with the domain shift using DG techniques. 
Thus, we first exploited a method which has been recently proposed to operate in this context, called Relative Norm Alignment (RNA) \cite{planamente2022domain}.  
This methods consists of an \textit{audio-visual domain alignment} at feature-level through the minimization of a cross-modal loss function ($\mathcal{L}_{RNA}$). The latter aims at minimizing the \textit{mean-feature-norm distance} between the audio and visual features norms among all the source domains, 
and it is defined as
\begin{equation}\label{formula:rna_1}
    \mathcal{L}_{RNA}=\left(\frac{\EX[h(X^v)]}{\EX[h(X^a)]} - 1\right)^2,
\end{equation}
where $h(x^m_i)=({\lVert{ \cdot }\rVert}_2 \circ f^m)(x^m_i)$ indicates the $L_2$-norm of the features $f^m$ of the $m$-th modality, $\EX[h(X^m)]=\frac{1}{N}\sum_{x^m_i \in \mathcal{X}^m}h(x^m_i)$ for the $m$-th modality and $N$ denotes the number of samples of the set $\mathcal{X}^m=\{x^m_1,...,x^m_N\}$.

Authors of \cite{planamente2022domain} proved that the norm unbalance between different modalities might cause the model to be biased towards the source domain that generate features with greater norm, thus causing wrong predictions.  
Contrarily, by simultaneously solving the problem of classification and relative norm alignment on different domains, the network extracts a shared knowledge between the different sources,  
resulting in a domain-agnostic model. 

In our submission to the EPIC-Kitchen UDA challenge, we extended the RNA-Net framework to the optical flow modality, in order to exploit the multiple sources available from the official training splits while showing the effectiveness of RNA loss in a multi-source DG setting. 

    


\subsection{Domain Adaptation}
The UDA techniques embedded into our pipeline can be divided in two main groups: \textit{feature}-level and \textit{classifier}-level. The first aims at aligning the distribution of source and target, and works at different levels of representation (frames- and video-level); the latter, instead, reduces the classifier's uncertainty on target data. 



\textbf{Multi-Level Adversarial Alignment}.

Following popular practices in unsupervised video domain adaption techniques, we integrate into our framework an adversarial approach \cite{videoda-chen2019temporal, munro2020multi}, consisting of an extension of the DANN \cite{grl-pmlr-v37-ganin15} standard UDA image-based method. We apply it at two different feature levels; frame- and video-level. It entails the introduction of two separate branches in our framework. Down-stream of said branches there are discriminators that try to distinguish the two domains (source and target). Contrarily, by maximising the corresponding discriminator losses, the network learns feature representations invariant to both domains.

\textbf{Attentive Entropy}. In order to reduce the uncertainty of the classifier on the target data, we minimize the attentive entropy loss proposed in \cite{videoda-chen2019temporal} as in \cite{plizzari2021polito}. This action minimizes the entropy, resulting in a refinement of the classifier adaptation. The term "attentive" refers to a loss re-weighting approach that prioritizes videos with low domain discrepancy by focusing on minimizing entropy for these videos. 



\subsection{Multi-Source Multi-Target Domain Adaptation}

The previous Epic Kitchen challenges \cite{ek19report, ek2020}, as well as the literature on unsupervised domain adaptation for first person action recognition \cite{Munro_2020_CVPR, planamente2022domain, planamente2021self, plizzari20212, plananamente2022test}, reveal a strong dependency of the models on the environment where the actions are recorded. This problem, known as ``environmental bias", causes a decrease in performance in occurrence of environment switches. As regards past action recognition challenges, we see this behavior by comparing performances of the models when tested on S1 (seen) and S2 (unseen). In the setting proposed in \cite{Munro_2020_CVPR}, similar behavior is observed, demonstrating the model's low generalization ability when tested on different kitchens.

The above considerations allow us to identify a secondary shift in this challenge, that occurs along with the temporal shift. Indeed, the training data are collected from different environments i.e. kitchens, thus introducing an environmental shift. As a result, we may rename the challenge setting \textit{Multi-Source Multi-Target Unsupervised Domain Adaptation}.


To deal with this new setting we propose a novel framework, which we call Multiple Spatio-Temporal Adversarial Alignment (MSTAA), combining Multiple Temporal Adversarial Alignment (MTAA) and Multiple Spatial Adversarial Alignment (MSAA). MTAA is obtained by adopting 2K domain adversarial branches (where K indicates the number of kitchens), aligning the source and the target distribution both at video- and frame-level for each kitchen. Instead, MSAA consists in adding another adversarial branch with a k-dimension discriminator in order to align the distribution of different kitchens and alleviate the environmental bias issue.

\input{table/EK_lead}

\subsection{Ensemble UDA losses}

For our final submission different models have been used in order to fully exploit the potentiality of popular video architectures. 
However, training individually each backbone with standard UDA protocols would result in independently adapted feature representations, which consequently vary between different streams. Our intuition is that this aspect could impact negatively the training process and the performance on target data. 
Indeed, since the domain adaption process acts on each architecture independently, naively training the backbones separately would yield mismatching prediction logits on target data, which, when combined, could increase the level of uncertainty of the model.
For this reason, we use the Min Entropy Consensus (MEC) loss, to impose a consistency constraint between feature representations from various models. Then, re-purposing the existing Complement Entropy (CENT) loss, we attempt to exploit the target data samples based on the assumption that there are some conditions in which it is easier to answer the question \textit{"Which classes does this action not belong to?"} rather than \textit{"Which class does this action belong to?"}.

\textbf{Min Entropy Consensus (MEC loss).} We extended the loss proposed in \cite{roy2019unsupervised} to encourage coherent predictions between different models. The resulting loss is defined as:
    \begin{equation}\label{eq:hna}
    \mathcal{L}_{MEC} = - \frac{1}{m}\sum_{i=1}^{m} \frac{1}{b} \max_{y \in \mathcal{Y} }{ \sum_{b} \log \textit{p}_b(y | x_{i}^{t})}
    \end{equation}
where $m$ is the cardinality of the batch size of the target set, $y$ is the predicted class, and $\log \textit{p}_b(y | x_{i}^{t})$ is the prediction probability of the $b$-th backbone network. The intuitive idea behind the proposed approach is to encourage different backbones to have a similar predictions.

\textbf{Complement Entropy (CENT).} The Complement Entropy (CENT) loss aims
at neutralizing the negative effects on the final prediction of clips whose logits present high degrees of uncertainty. 
It accomplishes this by “flattening" the predicted probabilities of “complement classes", i.e., all classes except the predicted one. As a result, when predictions are ensembled, the noise due to uncertainty on complement classes is reduced.
We refer to this loss as “complement entropy” objective, as it consists in maximizing the entropy for low-confident classes rather than minimizing it for the most confident one, as standard entropy minimization does. It is defined as:
\vspace{-0.3cm}
\begin{equation}
\begin{split}
    \mathcal{L}_{CENT} & =\frac{1}{N}\sum_{i=1}^N\mathcal{H}(\hat{y}_{i\Bar{c}}) 
    \\
    &=  -\frac{1}{N}\sum_{i=1}^N\sum_{j=1,j\neq p}^\mathcal{C}(\frac{\hat{y}_{ij}}{1-\hat{y}_{ip}}\log{\frac{\hat{y}_{ij}}{1-\hat{y}_{ip}}})%
    \\
    \end{split}
\end{equation}

where $N$ is the total number of samples in the batch, $\hat{y}_{ip}$ represents the predicted probability of the class $p$ with the higher score for the $i$-th sample, i.e., $\hat{y}_{ip}=max_j(\hat{y}_{ij})$, and $\mathcal{H}(\cdot)$ is the entropy function computed on the prediction of complement classes $\hat{y}_{i\Bar{c}}$ ($\Bar{c} \neq p$). The formulation is similar to the one in \cite{chen2019complement}, and we extend it to operate in an unsupervised fashion.
 

%% file: table/EK_lead.tex
\setlength\heavyrulewidth{0.31ex}

\begin{table*}[t]
\centering
\begin{adjustbox}{width=0.85\linewidth, margin=0ex 1ex 0ex 0ex}
\begin{tabular}{l|c|cccccc}
\toprule\noalign{\smallskip}
\multicolumn{8}{c}{\normalsize\textsc{Unsupervised Domain Adaptation Leaderboard}} \\
\noalign{\smallskip}
\cline{1-8}
\noalign{\smallskip}
  & Rank & \multicolumn{1}{c}{Verb Top-1} & Noun Top-1 & Action Top-1& Verb Top-5 & Noun Top-5 & Action Top-5\\ 
 \noalign{\smallskip} \hline
 
VI-I2R       & 1  & \textbf{57.89} & \underline{40.07} & \textbf{30.12} & \textbf{83.48} & \underline{64.19} & \textbf{48.10} \\ \hline 

Audio-Adaptive-CVPR2022  & 2 & 52.95 & \textbf{42.26} & \underline{28.06} & 80.03 & \textbf{67.51} & \underline{44.03}  \\ \hline 
 
\rowcolor[HTML]{AFDAAF}   plnet & 3 & \underline{55.51} & 35.86 & 25.25 & \underline{82.77} & 60.65 & 40.09  \\ \hline
 
CVPR2021-chengyi        & 4 & 53.16 & 34.86 & 25.00 & 80.74 & 59.30 & 40.75        \\ \hline
CVPR2021-M3EM & 5 & 53.29 & 35.64& 24.76 & 81.64 & 59.89 & 40.73        \\ \hline
CVPR2021-plnet & 6 &  55.22 & 34.83 & 24.71 & 81.93  & 60.48 & 41.41 \\ \hline
 EPIC\_TA3N \cite{damen2020rescaling}      & 8      & 46.91      & 27.69      & 18.95        & 72.70      & 50.72      & 30.53        \\ \hline
 EPIC\_TA3N\_SOURCE\_ONLY  \cite{damen2020rescaling} & 9 & 44.39      & 25.30      & 16.79        & 69.69      & 48.40      & 29.06   \\ 
\bottomrule
\end{tabular}
\end{adjustbox}
\caption{Leaderboard results of EPIC-Kitchens Unsupervised Domain Adaptation Challenge. The results obtained by the top-3 participants and the provided baseline methods are reported. \textbf{Bold:} highest result \underline{Underline}: second highest result; {\color{Green}{\textbf{Green:}}} our final submission. }
\label{leaderboard}
\end{table*}

\begin{table}[]
\begin{adjustbox}{width=1\linewidth, margin=0ex 0ex 0ex 0ex}
\begin{tabular}{l|ccc}
\toprule\noalign{\smallskip}
\multicolumn{4}{c}{\normalsize\textsc{Unsupervised Domain Adaptation}} \\
\noalign{\smallskip}
\cline{1-4}
 \noalign{\smallskip}  
  & \multicolumn{1}{c}{Verb} & Noun & Action \\ \hline \noalign{\smallskip}
{Ensemble} (E) \textit{Source Only} & 53.64 & 32.65 & 22.98 \\ \hline \noalign{\smallskip}
E-UDA & 53.88 & 33.10 & 23.22 \\ \hline \noalign{\smallskip}
E+MEC & 53.67 & 34.32 & 23.91  \\ \hline \noalign{\smallskip}
E+MEC+CENT & 54.20 & 33.92 & 23.99 \\ 
\hline \noalign{\smallskip}
E-SMR+MEC+CENT & \textbf{54.55} & 34.72 & 24.22 \\ 
\hline \noalign{\smallskip}
E-SMR+MEC+CENT+MTAA  & 54.09 & 33.72 & 23.77 \\ 
\hline \noalign{\smallskip}
E-SMR+MEC+CENT+MSTAA  & 54.01 & \textbf{34.82} & \textbf{24.24} \\ 
\bottomrule
\end{tabular}
\end{adjustbox}
\caption{Results on the EPIC-Kitchen validation set.}
\label{uda}
\end{table}

\begin{table}[]
    \begin{adjustbox}{width=1\columnwidth, margin=0ex 0ex 0ex 0ex}
\begin{tabular}{l|ccc}
\toprule\noalign{\smallskip}
\multicolumn{4}{c}{\normalsize\textsc{Domain Generalization}} \\
\noalign{\smallskip}
\cline{1-4}
\noalign{\smallskip}
  & \multicolumn{1}{c}{Target} & Verb Top-1 & Verb Top-5 \\ 
 \noalign{\smallskip} \hline \noalign{\smallskip}

Source Only & \xmark               & 44.39 & 69.69 \\ \hline \noalign{\smallskip}

EPIC\_TA3N  \cite{damen2020rescaling}      & \cmark               & 46.91 & 72.70 \\ \hline \noalign{\smallskip}

RNA-Net \cite{planamente2022domain}        & \xmark              & \underline{47.96} & \underline{79.54} \\ \hline
\noalign{\smallskip}
EPIC\_TA3N+RNA-Net    & \cmark               & \textbf{50.40} & \textbf{80.47} \\ 
\bottomrule
\end{tabular}
\end{adjustbox}
\caption{Results on the EPIC-Kitchen test set.}
\label{dg}
\end{table}

%% file: root/backbone.tex
\section{Framework}
In this section, we describe the architectures of the feature extractors used to produce suitable multi-modal video embeddings, and the fusion stategies adopted to combine them. Finally, we deepen the analysis describing the hyper-parameters used for the training.

\subsection{Architecture}
\setlength\heavyrulewidth{0.31ex}

\textbf{Backbone.} 
For our submission, we adopted three different network configurations.
In the first one, corresponding to the RNA-Net framework in \cite{planamente2022domain}, 
we used the Inflated 3D ConvNet (I3D), pre-trained on Kinetics \cite{carreira2017quo}, for RGB and Flow streams, and a BN-Inception model \cite{ioffe2015batch} pre-trained on ImageNet \cite{imageNet} for the auditory information. 
Each feature extractor produces a 1024-dimensional representation which is fed to an action classifier. 
In the second configuration, we used BN-Inception models for all the three streams, using pre-extracted features from a TBN \cite{munro2020multi} model trained on EPIC-Kitchens-55. In the last configurations, we used standard ResNet-50 architectures \cite{he2016deep} equipped with the Temporal Shift Module ~\cite{lin2019tsm}  pre-trained on EPIC-Kitchens-55 \footnote{\url{https://github.com/epic-kitchens/epic-kitchens-55-action-models}}. 

\textbf{Multi-modal fusion strategies.}
In all the above mentioned configurations, each modality is processed by its own backbone, and the corresponding extracted representations are then fused following different strategies.
For RNA-Net, we followed a standard late fusion strategy, consisting in averaging the final score predictions obtained from two different fully-connected layers (verb, noun) from each modality. In the other configurations, we adopted the recent mid-fusion strategy, called Semantic Mutual Refinement sub-module (SMR), proposed in \cite{yang2021epic}, to generate a common frame-embedding among the modalities. Then, using temporal pooling, we obtain a final video-embedding that is sent to the verb and noun classifiers. 

\setlength{\tabcolsep}{10pt}

\begin{table}[t]
\centering
\begin{adjustbox}{width=\columnwidth, margin=0ex 1ex 0ex 0ex}
\begin{tabular}{|c|c|c|c|c|}
\hline
$\lambda_{RNA}$ & $\lambda_{CENT}$  & $\lambda_{MEC}$ & $\gamma$ & $\beta$         \\ \hline
1             & 0.31       & 0.22    & 0.003 & 0.75, 0.75, 0.75 \\ \hline
\end{tabular}
\end{adjustbox}
\caption{UDA losses hyper-parameters used during training.} 
\label{params}
\end{table}

\subsection{Implementation Details}
We trained I3D and BN-Inception models with SGD optimizer, with an initial learning rate of 0.001, dropout 0.7, and using a batch size of 128, following \cite{planamente2022domain}. Instead, when using pre-extracted features from ResNet50 or BNInception, we trained the SMR modules on top of them for 45 epochs with an initial learning rate of 0.03, decayed after epochs 25 and 35 by a factor of 0.1. We used a batch size of 128 with SGD optimizer. In Table \ref{params} we report the other hyper-parameter used. Specifically, we indicate with $\lambda_{RNA}$, $\lambda_{CENT}$ and $\lambda_{MEC}$ the weights of RNA, CENT and MEC losses respectively. In addition, we report the values used to weight the attentive entropy loss ($\gamma$) and the domain losses at different levels ($\beta$) for MSTAA. 

%% file: root/Exp.tex
\section{Results and Discussion}


In Table \ref{leaderboard} we report our best performing model on the target test, achieving the \textbf{2st} position on ‘verb', and the \textbf{3rd} on ‘noun' and ‘action'. Meanwhile, in Tables \ref{uda} and \ref{dg} we show an ablation of the proposed UDA and DG methods described in section \ref{OurApp}.

\textit{How well do DG approaches perform? } The results in Table \ref{dg} are obtained under the multi-source DG setting, when target data are not available during training. Noticeably, RNA outperforms the baseline Source Only by up to $3\%$ on Top-1 and $10\%$ on Top-5, highlighting the importance of using ad-hoc alignment techniques to deal with multiple sources in order to effectively extract a domain-agnostic model. Moreover, it outperforms the recent UDA technique TA$^3$N \cite{videoda-chen2019temporal} without accessing target data. 
Interestingly, when combined with EPIC\_TA3N, it further improves performance, proving the complementarity of RNA to other existing UDA approaches. 

In Table \ref{uda} it can be seen how the proposed UDA approaches improve Top-1 accuracy on all categories by up to $1\%$. Although using an additional adversarial branch for each kitchen does not appear to provide a significant improvement on the validation set, it increases the top-1 action accuracy on the test set, allowing us to obtain the third position in the challenge. Without MSTAA, the accuracy on the action top-1 reaches just 24.83\%. 
This outcome was predictable given that the validation set is populated with a different set of kitchens than the test set, whereas the kitchens in the test set are the same as those used for the target and source training. This aspect confirms the \textit{Multi-Source Multi-Target Unsupervised Domain Adaptation} setting and the presence of two different shifts, the \textit{temporal} shift (2018-2020) and the \textit{environmental} shift (among the kitchens).

{\small 
\textbf{Acknowledgements. }
This work was supported by the CINI Consortium through the VIDESEC project and by the Italian Ministry of University and Research under the DM1061. The research herein was carried out using the IIT HPC infrastructure. 
}